\documentclass[letterpaper]{article}
\usepackage{uai2020}
\usepackage[margin=1in]{geometry}

\usepackage{times}
\usepackage[utf8]{inputenc} 
\usepackage[T1]{fontenc}    
\usepackage[colorlinks,citecolor=blue,urlcolor=blue,linkcolor=blue,bookmarks=false]{hyperref}       
\usepackage{url}            
\usepackage{booktabs}       
\usepackage{amsfonts}       
\usepackage{nicefrac}       
\usepackage{microtype}      
\usepackage{xcolor}
\usepackage{sidecap}
\usepackage{wrapfig}
\usepackage[numbers,sort&compress]{natbib}

\usepackage{graphicx} 
\usepackage{booktabs} 
\usepackage{array} 
\usepackage{mathtools} 
\usepackage{amsthm, amssymb, amscd, amsfonts} 
\usepackage{amsmath}
\usepackage{bbm}
\usepackage[ampersand]{easylist} 
\usepackage{multirow} 
\usepackage{booktabs} 
\usepackage{bm} 
\usepackage{dcolumn} 
\usepackage{caption}
\usepackage{subcaption} 
\usepackage{epstopdf} 
\usepackage[export]{adjustbox}
\usepackage{enumitem}
\usepackage{algorithm} 
\usepackage{algorithmic} 
\usepackage{xspace}

\newtheorem{example}{Example}

\DeclareMathOperator*{\argmin}{arg\,min\,}
\DeclareMathOperator*{\argmax}{arg\,max\,}

\newcommand{\R}{\ensuremath{\mathbb{R}}}

\newcommand{\loss}{\mathcal{L}}

\newcommand{\nvar}{d}
\newcommand{\ndim}{\nvar}

\newcommand{\ii}{i}
\renewcommand{\ij}{j}

\newcommand{\x}{x}
\newcommand{\xvec}{\bm{\x}}

\newcommand{\y}{y}
\newcommand{\yvec}{\bm{\y}}

\newcommand{\z}{z}
\newcommand{\zvec}{\bm{\z}}

\renewcommand{\u}{u}
\newcommand{\uvec}{\bm{\u}}

\renewcommand{\v}{v}
\newcommand{\vvec}{\bm{\v}}

\newcommand{\I}{\mathbf{I}}

\newcommand{\E}{\mathbb{E}}

\theoremstyle{definition}

\theoremstyle{definition}
\newtheorem*{ex*}{Example}

\renewcommand{\cite}[2][]{\citep[#1]{#2}}
\newcommand{\one}{\mathbbm{1}}

\newcommand{\curve}{\phi}

\newcommand{\nk}{k} %
 
\title{Automated Dependence Plots}

\author{ {\bf David I. Inouye
} \\ 
Purdue University\\
dinouye@purdue.edu \\
\And
{\bf Liu Leqi, Joon Sik Kim}  \\
Carnegie Mellon University\\
\{leqil,joonsikk\}@cs.cmu.edu\\
\And
{\bf Bryon Aragam}   \\
University of Chicago\\
bryon@chicagobooth.edu
\And
{\bf Pradeep Ravikumar}   \\
Carnegie Mellon University\\
pradeepr@cs.cmu.edu \\
}

\begin{document}

\maketitle

\begin{abstract}
In practical applications of machine learning, it is necessary to look beyond standard metrics such as test accuracy in order to validate various qualitative properties of a model. Partial dependence plots (PDP), including instance-specific PDPs (i.e., ICE plots), have been widely used as a visual tool to understand or validate a model. Yet, current PDPs suffer from two main drawbacks: (1) a user must manually sort or select interesting plots, and (2) PDPs are usually limited to plots along a single feature. To address these drawbacks, we formalize a method for automating the selection of interesting PDPs and extend PDPs beyond showing single features to show the model response along arbitrary directions, for example in raw feature space or a latent space arising from some generative model. We demonstrate the usefulness of our automated dependence plots (ADP) across multiple use-cases and datasets including model selection, bias detection, understanding out-of-sample behavior, and exploring the latent space of a generative model. The code is available at \url{https://github.com/davidinouye/adp}. 
\end{abstract}

\section{INTRODUCTION}
\label{sec:intro}

Modern applications of machine learning (ML) involve a complex web of social, political, and regulatory issues as well as standard technical issues such as covariate shift or training dataset bias.
Although the most common validation metric is test set accuracy, the aforementioned issues cannot be resolved by test set accuracy alone---and sometimes these issues directly oppose high test set accuracy (e.g., privacy concerns).
Especially in certain applications such as autonomous cars or automated loan approval, practitioners have become concerned with unexpected and incorrect model behaviors, such as behaviors for data points that were not seen during training or testing (e.g., classifying a person as part of the road or approving a large fraudulent loan).
Thus, average-based validation may be insufficient to validate a model.
Rather, given the difficulty of specifying expected model behavior \emph{a priori}, qualitative methods for validating a model and highlighting the most interesting or unusual model behaviors beyond average-based methods are needed. 

One popular approach for qualitatively validating or understanding the effect of a particular feature on the model response is a partial dependence plot (PDP) \citep{Friedman2001}.
A PDP plots the average model response across one feature marginalized over the other features and thus gives a global view of the feature effect on the model response.
Because PDPs take an average over some features, they may not be as effective at showing unusual behaviors that may be important for certain applications as noted above.
To address this gap, instance-specific PDPs such as Individual Conditional Expectation (ICE) plots \citep{Goldstein2015} could be used to show the PDP plot for a single target instance rather than averaged over a set of instances.
While PDPs were invented many years ago, they have continued to be widely used for understanding model behavior because they are simple to interpret (e.g., \citep{wexler2019if}).

In order to investigate unusual behaviors for safety-critical applications, practitioners must manually inspect $O(nd)$ instance-specific PDPs, where $n$ is the number of samples they want to investigate and $d$ is the number of features.
This manual inspection is prohibitive, even for moderate $n$ and $d$.
Additionally, because PDPs only consider a single feature (i.e., axis-aligned directions), they can miss important interactions between features which may be critical in certain applications.
One could use 2D PDP heatmaps, however, this would increase the number of plots from $O(nd)$ to $O(nd^2)$.
Thus, despite widespread use, current PDPs suffer from two main drawbacks: (1) a user must manually select interesting plots, and (2) PDPs are usually limited to plots along a single feature.

\begin{figure}[!ht]
    \centering
    \includegraphics[width=\linewidth]{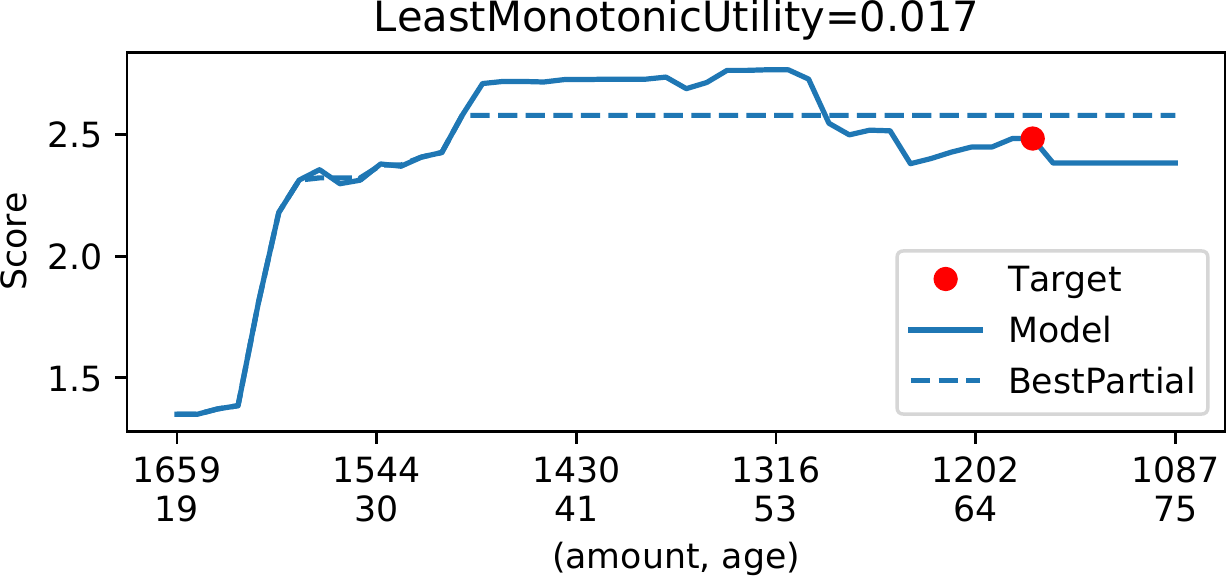}
    \vspace{-1.5em}
    \caption{
    The above example highlights the two key innovations of our proposed automated dependence plots: (1) Finding interesting plots by formalizing and optimizing plot utility measures---in this case, the utility is based on the non-monotonicity of the curve. (2) Optimizing over directions that change multiple features rather than only varying a single feature as in standard PDP plots (as indicated on the $x$-axis).
    For the target loan application (designated by the red point),
    the directional dependence plot (solid line) shows the change in model scores ($y$-axis) along a direction in which two numeric features (amount and age) are varied.
    The plot was optimized so that the model response along the plot is the least monotonic (for comparison, the best monotone regression line is shown via the dotted line;
    see Sec~\ref{sec:utility} for more details).
    }
    \vspace{-1em}
    \label{fig:main-illustration}
\end{figure}

To address these drawbacks, we formalize a method to automate the selection of interesting PDP plots and to extend PDPs to \emph{directional} dependence plots, which show the model response in more general directions---either in raw feature space for tabular data or in a latent space for, e.g., visual or textual data.
An illustrative example of such a directional dependence plot is given in \autoref{fig:main-illustration}. Here, multiple features are being changed and the specific plot is optimized to show the least monotonic direction. This optimization, which is one of our key contributions, is how we automate the selection of ``interesting'' or relevant plots.

We summarize our contributions as follows:
\begin{enumerate}[itemsep=.3mm]
    \item We formalize the concept of interestingness or unexpectedness of dependence plots by defining two classes of plot utility functions that have multiple instantiations including utilities to compare two models and validate (or invalidate) certain properties such as linearity, monotonicity, and Lipschitz continuity.
    \item We generalize PDPs beyond axis-aligned directions to consider the model response along sparse linear directions.
    In particular, we optimize a specified utility measure over sparse linear directions using a greedy coordinate pairs algorithm.
    For tabular data where the features are inherently interpretable, we optimize for a sparse linear direction in the raw feature space.
    For rich data such as images or text, we propose to find sparse linear directions in a latent representation space (e.g., via a VAE) but show the corresponding images along this direction in the original raw feature space.
    \item We demonstrate the usefulness of the resulting automated dependence plots (ADPs) across multiple use-cases and datasets including model selection, bias detection, understanding out-of-sample behavior, and exploring the latent space of a generative model. 
\end{enumerate}

\paragraph{Related work.}

The importance of safety and robustness in ML is now well-acknowledged \citep{varshney2017safety,saria2019tutorial}, especially given the current dynamic regulatory environment surrounding the use of AI in production \citep{hadfield2019safeml,wachter2017counterfactual}.
A popular approach to auditing and checking models before deployment is ``explaining'' a black-box model post-hoc.
Both early and recent work in explainable ML rely on local approximations (e.g., \cite{Ribeiro2016, guidotti2018local}).
Other recent papers have studied the influence of training samples or features \citep{datta2016algorithmic,dhurandhar2018explanations}.
These have been extended to subsets of informative features \citep{Lundberg2017a,datta2016algorithmic,chen2018learning} to explain a prediction.
Other approaches employ influence functions \cite{koh2017understanding} and prototype selection \cite{yeh2018representer,zhang2018interpreting}. 
A popular class of approaches take the perspective of local approximations, such as linear approximations \citep{Ribeiro2016,Lundberg2017a}, and saliency maps \citep{simonyan2013deep, sundararajan2017axiomatic, smilkov2017smoothgrad}.
A crucial caveat with such approximation-based methods is that the quality of the approximation is often unknown, and is typically only valid in small neighborhoods of an instance (although we note recent work on global approximations \cite{guo2018explaining}).
In contrast to these previous feature selection methods, our approach leverages a utility measure to select features or directions.

\section{PLOT UTILITY MEASURES}
\label{sec:utility}

In this section, we define two classes of utility measures for generic dependence plots including PDP plots, ICE plots, and our generalization to directional dependence plots described in the next section.
The intuition is that the plot utility measure quantifies the ``interestingness'' or relevance of a particular feature or direction.
As we will discuss in more detail in Sec.~\ref{sec:opt}, it is implicit here that we wish to \emph{maximize} the utility (cf. \eqref{eqn:instance-optimization}).

\paragraph{Notation.}
We assume that the input space is $\mathcal{X}=\R^{d}$---we could consider categorical variables by showing bar charts as in categorical PDP plots but for simplicity, we will focus on continuous features.

For instance-specific plots, we will denote the target instance as $\xvec_0 \in \R^d$.
Let $f \colon \R^d \to \R$ be a black-box function, i.e., we can query $f$ to obtain pairs $(\xvec, f(\xvec))$.
We do not assume that $f$ is differentiable or even continuous in order to allow non-differentiable models such as random forests.
We will denote a plot by it's corresponding univariate function $\tilde{f}(t) \colon \R \to \R$, where we visualize this function for some bounded interval, i.e., for $t \in [a, b]$---the bounds for standard PDP plots are usually based on the minimum and maximum values along each feature but we will generalize this for directional plots in the next section.
For other multivariate functions $g(\xvec)$, we will denote similarly the corresponding univariate function as $\tilde{g}(t)$.
As an example, for PDP $\tilde{f}_i(t) = \E[f(\dots, x_{i-1}, t, x_{i+1},\cdots)]$ where the expectation is an empirical expectation with respect to some dataset (often a training dataset) and $t$ ranges over the minimum and maximum of the $i$-th feature.
For instance-specific PDP (ICE), $\tilde{f}_i(t) = f(\dots, x_{i-1}, t, x_{i+1},\cdots)$ where the input $\xvec$ is fixed to a target point $\xvec_0$ except for the $i$-th feature---this can be seen as a PDP plot where the dataset is a single point.
We will denote $U(\tilde{f}, a, b)$ as a plot utility function where we usually suppress the dependence on the bounds $a$ and $b$ for simplicity and merely denote the utility as $U(\tilde{f})$.

In the next subsections, we carefully develop and define two general classes of plot utility measures.
Both classes of utilities compare to another plot, which we will show as a dotted line as can be seen in \autoref{fig:main-illustration}.
This helps give the reader an interpretable reference for the utility measure itself---e.g., showing the best bounded Lipschitz approximation to the plot.
Of course, we do not claim that our proposed set of utility measures is exhaustive, and indeed, this paper will hopefully lead to more creative and broader classes of utility measures.

\subsection{MODEL CONTRAST UTILITY MEASURES} 
A natural way to measure the utility of a plot with respect to one model is to contrast it to the same plot based on a different model. 
Given another multivariate model $g_{\xvec_{0}}(\xvec) \colon \R^d \to\R$, which could be defined with respect to a target instance, we define the contrast utility measure:
\begin{align}
\label{eq:def:Uglobal}
\textstyle{U_{\text{c}}(\tilde{f}) = \int_a^b \loss(\, \tilde{f}(t), \,\, \tilde{g}_{\xvec_0}(t) \,) dt} \, ,
\end{align}
where $\loss$ is a loss function, e.g., squared or absolute loss, and $\tilde{g}_{\xvec_0}(t)$ is the corresponding plot function with respect to $g_{\xvec_0}(\xvec)$.
By maximizing $U_{\text{c}}(\tilde{f})$, we can find the plot $\tilde{f}$ that differs the \emph{most} from the baseline model $\tilde{g}_{\xvec_0}(t)$.
The baseline could be another simple model like logistic regression or some local approximation of the model around the target instance such as explanations produced by LIME \citep{Ribeiro2016}---this would provide a way to critique local approximation methods and show where and how they differ from the true model.

\begin{example}[Variance of Plot]
While the PDP paper \citep{Friedman2001} did not propose ways for sorting or selecting interesting plots, a commonly used method to sort PDP curves is by the variance of the plot. 
The utility of a plot based on variance can be seen as special case of the model contrast utility where $\tilde{g}(\xvec) = c = \E[\tilde{f}(s)] = 1/(b-a)\int_a^b \tilde{f}(s) ds$---i.e., an expectation of $\tilde{f}$ with respect to a uniform distribution between $a$ and $b$---and the loss function is squared error.
We will use this as the default sorting mechanism for comparison in our experiments because it is the simplest and seems to be common in practice.
\end{example}

\begin{example}[Contrast with Constant Model]
Sometimes we may want to create a contrast model that is dependent on the target point $\xvec_0$.
The simplest case is where the contrast model is a constant fixed to the original prediction value, i.e., $g_{\xvec_0}(\xvec) = c_{\xvec_0} = f(\xvec_0)$
Note that in this case the comparison function depends on $\xvec_0$.
This contrast to a constant model can find directions that deviate the most from the prediction; this implicitly finds plots that are not flat and significantly affect the prediction value.
\end{example}

\begin{example}[Contrast with Validated Linear Model]
Suppose an organization has already deployed a carefully validated linear model---i.e., the linear parameters were carefully checked by domain experts to make sure the model behaves reasonably with respect to all features.
The organization would like to improve the model's performance by using a new model, but wants to see how the new model compares to their carefully validated linear model to see where it differs the most. 
In this case, the organization could let the contrast model be their previous model, i.e., $g_{\xvec_0}(\xvec)=g_{\textnormal{Linear}}(\xvec)$ where $g$ does not depend on the target point $\xvec_0$.
\end{example}

\begin{example}[Contrast Random Forest and DNN]
An organization may want to compare two different model classes such as random forests and deep neural networks (DNN) to diagnose if there are significant differences in these model classes or if they are relatively similar.
In this case, the contrast model $g(\xvec)$ would be the random forest or DNN.
\end{example}

\begin{figure}[!ht]
    \centering
    \includegraphics[width=\linewidth]{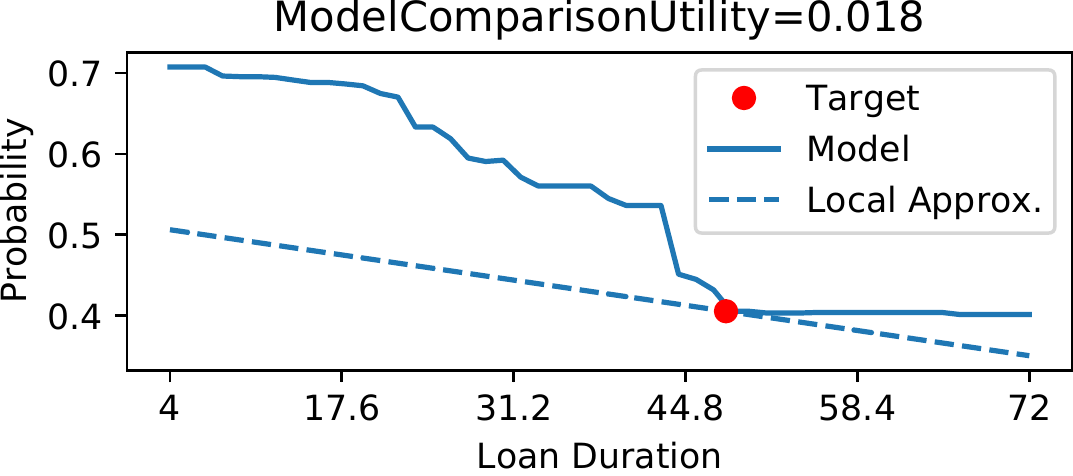}
    \vspace{-2em}
    \caption{This plot illustrates using the model contrast utility where $g_{\xvec_0}$ (dotted line) is an explanation model based on the gradient similar to the local linear explanation models in LIME \citep{Ribeiro2016}.
    Notice how it shows where the approximation may be appropriate (duration > 46) and where it might be far from the true model (duration < 46).
    }
    \vspace{-1em}
    \label{fig:lime-contrast}
\end{figure}

\begin{example}[Contrast with local approximations used for explanations]
We can also compare the true model with explanation methods based on local approximation such as LIME \citep{Ribeiro2016} or gradient-based explanation methods \citep{sundararajan2017axiomatic, shrikumar2017learning, smilkov2017smoothgrad}.
We can simply use the local approximation to the model centered at the target point as the contrast model,i.e., $g_{\xvec_0}(\xvec) = \hat{f}_{\xvec_0}(\xvec)$, where $\hat{f}_{\xvec_0}$ is the local approximation centered around $\xvec_0$.
Thus, the found diagnostic curve will show the maximum deviation of the true model from the local approximation model being used for an explanation.
Importantly, this allows our diagnostic method to assess the goodness of local approximation explanation methods showing when they are reasonable and when they may fail; see \autoref{fig:lime-contrast}.
\end{example}

\subsection{FUNCTIONAL PROPERTY (IN)VALIDATION UTILITY MEASURES}
In many contexts, a user may be more interested in validating (or invalidating) certain functional properties of a model, such as monotonicity or smoothness.
For example, if it is expected that a model should be increasing with respect to a feature (e.g., income in a loan scoring model), then we'd like to check that this property holds true (at least approximately).
Let $\mathcal{H}$ be a class of univariate functions $\tilde{h}:\R \to \R$ that represents a property that encodes acceptable or expected behaviors.
To measure deviation of a plot from this class of functions, take the minimum expected loss over all $h\in\mathcal{H}$:
\begin{align}
\begin{aligned}
    &U_{\text{p}}(\tilde{f}) 
    = \min_{\tilde{h} \in \mathcal{H}} \int_a^b \loss(\tilde{f}(t), \tilde{h}(t)) dt
\end{aligned}
    \label{eqn:utility-partial}
\end{align}
where as usual, $\loss$ is a loss function.
The minimization in \eqref{eqn:utility-partial} is a univariate regression problem that can be solved using standard techniques. 
This utility will find dependence plots that maximally or minimally violate the functional properties encoded by $\mathcal{H}$.

\begin{example}[Linearity (in)validation via linear regression]
A user might want to view the plots that are the least linear to see if there is some unusual plots that may need further investigation.
For this example, the class of functions would merely be linear functions, i.e., $\mathcal{H}$ is the set of univariate linear functions.
This problem could be solved easily using standard linear regression methods.
\end{example}

\begin{example}[Monotonicity (in)validation via isotonic regression]
In many applications, it may be known that the model output should behave simply with respect to certain features. For example, one might expect that the score is monotonic in terms of income in a loan scoring model. 
In this case, the class of functions should be the set of monotonic functions, i.e., $\mathcal{H}$ is the set of all monotonic functions.
The resulting problem can be efficiently solved using isotonic regression \citep{best1990active}---and this is what we do in our experiments; see \autoref{fig:loan-monotonic-comparison} for an example of validating (or invalidating) the monotonic property.
\end{example}

\begin{figure}[!ht]
    \centering
    \vspace{-1em}
    \includegraphics[width=0.9\linewidth]{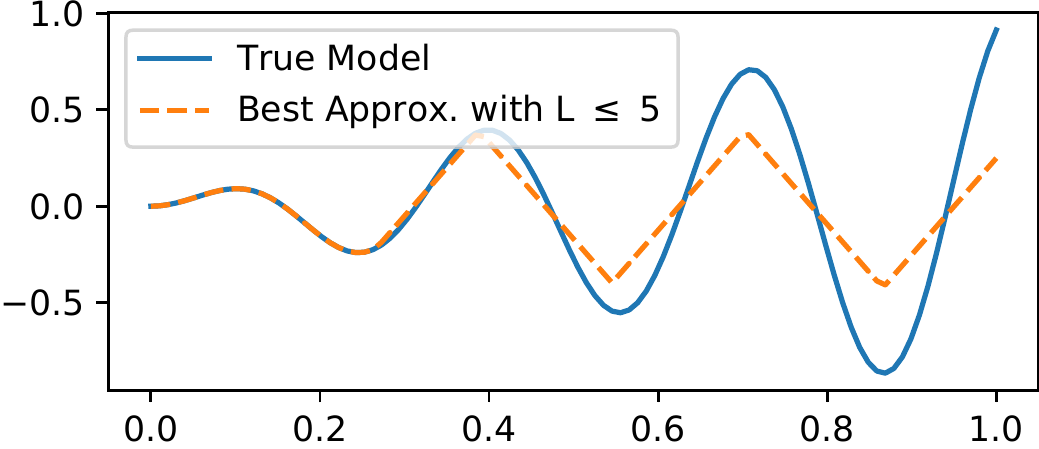}
    \vspace{-0.5em}
    \caption{Example of contrast utility function when the function class is Lipschitz continuous with $L \le 5$. }
    \label{fig:best-lipschitz-illustration}
\end{figure}

\begin{example}[Lipschitz-boundedness (in)validation via constrained least squares] 
Another property that an organization might want to validate is whether the function has a small Lipschitz constant along the curve.
Formally, they may wish to check if the following condition holds:
\begin{align}
    \left|\frac{\tilde{f}(t_2) - \tilde{f}(t_1)}{t_2 - t_1}\right| \leq L \,, \quad \forall t_1, t_2 \in [a, b]
\end{align}
where $L$ is a fixed Lipschitz constant.
Thus, the corresponding class of functions $\mathcal{H}_L$ is the set of Lipschitz continuous functions with a Lipschitz constant less than $L$.
In practice, we can solve this problem via constrained least squares optimization---similar to isotonic regression (details in supplement).
An example of using Lipschitz bounded functions for $\mathcal{H}$ can be seen in \autoref{fig:best-lipschitz-illustration}.
This utility will find the curve that maximally violates the bounded Lipschitz condition; this curve may also be useful in finding where the model is particularly sensitive to changes in the input since the derivative along the curve will be larger than the Lipschitz constant.
\end{example}

\section{DIRECTIONAL DEPENDENCE PLOTS}
\label{sec:param}

In this section, we formally define a directional dependence plot, which generalizes classical PDPs and instance-specific PDPs.
First, recall that a PDP $\tilde{f}_i$ varies the $i$-th feature and averages over the other features in a dataset, i.e., $\tilde{f}_i(t) = \E_{\xvec}[f(\xvec + t\mathbf{e}_i)]$, where $\mathbf{e}_i$ is the $i$-th standard basis vector.
In this form of $\tilde{f}_i$, the generalization to directions is quite straightforward as we can replace $\mathbf{e}_i$ by $\vvec$ where $\vvec$ is an arbitrary unit vector: $\tilde{f}_{\vvec}(t) = \E_{\xvec}[f(\xvec + t\vvec)]$.
The instance-specific dependence plot with respect to $\xvec_0$ is:
\begin{align}
    \tilde{f}_{\vvec, \xvec_0}(t) = f(\xvec_0 + t\vvec) \, .
    \label{eqn:}
\end{align}
In future sections, we will often suppress the dependence on $\vvec$ or $\xvec_0$ if it is understood from the context.
Because this paper focuses on finding unusual or interesting plots for the purposes of model validation especially in safety-critical applications, we will focus on the local instance-specific directional plots in our experiments but these ideas can clearly be applied to ``global'' directional plots.
We note that both for interpretability and computational reasons, we will usually assume that $\vvec$ is sparse or low-dimensional in the following sections.
In the next sections, we will develop generalizations for directional plots in latent spaces, parameter spaces of transformations, and other general considerations for directional plots.

\paragraph{Directions in a latent space.}
For rich data such as images, directions in the original input space (e.g., raw pixels) may not be very informative.
In these cases, it may be more intuitive to move along directions in a latent space, such as one arising from a generative model $G: \R^{\bar{d}} \mapsto \R^d$, that generates input vectors in $\R^d$ given latent vectors in $\R^{\bar{d}}$, where $\bar{d}$ is usually smaller than $d$ (e.g., a low dimensional representation).
Examples of such models include VAEs \citep{Kingma2014,  Rezende2014} and deep generative models via normalizing flows \cite{Germain2015,Papamakarios2017,Oliva2018a,Dinh2017,inouye2018deep}.
We then optimize for directions in the latent space of the generative model $G(\cdot)$ rather than the raw input space itself.
This allows us to define directions in the latent space that can correspond to arbitrary curves in the latent space:
\begin{align}
    \tilde{f}_{G, \xvec_0}(t) = f( G( \zvec_0 + t\vvec ) ) \, ,
\end{align}
where $\zvec_0 = G^{-1}(\xvec_0)$ and $G^{-1}(\xvec_0)$ is an (approximate) input to the generative model that would have generated $\xvec_0$, i.e., $G(G^{-1}(\xvec_0)) \approx \xvec_0$.
For VAEs \citep{Kingma2014, Rezende2014}, the decoder network acts as $G$ and the encoder network acts as an approximate $G^{-1}$.
For normalizing flow-based models, both $G(\zvec)$ and $G^{-1}(\xvec_0)$ can be computed exactly by construction \cite{Germain2015,Papamakarios2017,Oliva2018a,Dinh2017,inouye2018deep}.

\paragraph{Directions in the parameter space of known transformations.}
\renewcommand{\nk}{\ell} 
In certain domains, there are natural classes of transformations that are semantically meaningful. 
Examples include adding white noise to an audio file or blurring an image, removing semantic content like one letter from a stop sign, or changing the style of an image using style transfer~\citep{johnson2016perceptual}.
We can consider directions in the parameter space of a set of these known transformations, which could be arbitrarily complex and non-linear.
We will denote each transformation as 
$\lambda_v \colon \R^\ndim \to \R^\ndim$ with parameter $v \in [0,1]$, where $\lambda_0$ corresponds to the identity transformation.
For example, if $\lambda_v$ a rotation operator, then $v$ would represent the angle of rotation where $v=0$ is 0 degrees and $v=1$ is 180 degrees.
Given an ordered set of $\nk$ different transformations denoted by $(\lambda_v^{(1)}, \lambda_v^{(2)}, \dots, \lambda_v^{(\nk)} )$,
we can define a plot function using a composition of these simpler transformations:
\begin{align}
\begin{aligned}
    \tilde{f}_{\vvec, \xvec_0}(t) &= f(\Lambda_{\xvec_0}(\vvec t)) \\
    \Lambda_{\xvec_0}(\vvec) &\triangleq \lambda_{v_\nk}^{(\nk)}(\dots \lambda_{v_2}^{(2)}( \lambda_{v_1}^{(1)}(\xvec_0))\dots)
\end{aligned}
\end{align}
where $\vvec \in [0,1]^\nk$ is the optimization parameter. 
Thus, directions in the parameter space correspond to non-linear transformation curves in the original input space.
We note that the transformations can be arbitrarily non-linear and complex---even deep neural networks.
\renewcommand{\nk}{k} 

\paragraph{Realistic plot bounds.}
In some contexts, it is desirable to explicitly audit the behavior of $f$ off its training data manifold. This is helpful for detecting unwanted bias and checking robustness to rare, confusing, or adversarial examples---especially in safety critical applications such as autonomous vehicles.
In other contexts, we may wish to ensure that the selected plots show realistic combinations of variables---that is, that they stay within the bounds of the training data distribution.
For example, it may not be useful to show an example of a high income person who also receives social security benefits.
Fortunately, it is not hard to enforce this constraint: 
First, bound the endpoints of the plot using a box constraint based on the minimum and maximum of each feature---this is all that classical PDP plots consider.
However, this simple bounding may not be reasonable when we can move along multiple features.
Thus, we can also train a density model on the training data to account for feature dependencies and further restrict the bounds of the plot to only lie in regions with a density value above a threshold.
In our experiments, we use a simple multivariate Gaussian density estimate to create the boundaries of our plots but more complex deep density estimators could be used (e.g.,  \cite{Papamakarios2017,Oliva2018a,Dinh2017,inouye2018deep}).

\paragraph{Visualizing directional dependence plots.}
In typical PDP plots, the feature values are shown on the horizontal axis---yielding direct interpretability if the feature itself is interpretable.
For directional plots, we show the set of feature values that are changing that correspond to each value of $t$ (see the $x$-axis of \autoref{fig:main-illustration}).
For example, if both age and duration of a loan application are changing across a directional plot, we show pairs of age and duration along the horizontal axis.
For image-based examples, we show the image in the original feature space corresponding to different $t$ values to maintain interpretability.

\section{OPTIMIZATION OF PLOT UTILITY}
\label{sec:opt}
So far we have discussed: 
1) how to measure the plot utility and 2) how to generalize beyond axis-aligned directions to arbitrary directions.
With these pieces in place, the objective is to find directional dependence plots that are simultaneously interpretable and interesting.
To do this, we restrict the directions under consideration (usually to sparse directions) and optimize for the direction that shows the best (either highest or lowest) utility:
\begin{align}
    \vvec^* = \argmax_{\vvec} U(\tilde{f}_{\vvec, \xvec_0}) \, ,
    \label{eqn:instance-optimization}
\end{align}
where $\vvec$ is restricted to some interpretable subset (e.g., sparse vectors).
This optimization problem can also be optimized over a set of target instances as follows:
\begin{align}
    \vvec^*, \xvec_0^* = \argmax_{\vvec, \xvec_0} U(\tilde{f}_{\vvec, \xvec_0}) \, ,
    \label{eqn:main-optimization}
\end{align}
where $\xvec$ is one instance in a specified dataset (e.g., a random sample of the training dataset).
This finds both a target data point and a direction where $f$ exhibits unusual or interesting behavior defined by the utility $U$, thus enabling users to see the worst case behavior of the model.
Note that the dataset used for optimization may not be part of the training data because it may be new data where no class labels are given.

Since $f$ is assumed to be an arbitrary black-box, where only function evaluations are possible, the optimization problem even for a single target instance in \eqref{eqn:instance-optimization} is usually nonconvex and nonsmooth.
This model-agnostic setup generalizes to many interesting settings including ones in which the model is private, distributed or nondifferentiable (e.g., boosted trees) such that exact gradient information would be impossible or difficult to compute.
Given these restrictions, we must resort to zeroth-order optimization.
While we could use general zeroth-order optimization techniques, we require that the directions are sparse so that the resulting directional plots are interpretable.
Thus for both computational and practical reasons, we propose a greedy optimization scheme called \emph{greedy coordinate pairs} (GCP) that adds non-zero coordinates in a greedy fashion, as outlined in Algorithm~\ref{alg:greedy-ascent}.
We initialize this algorithm by finding the single feature with the highest utility---this is the same as computing the best axis-aligned direction for standard PDP plots.

\renewcommand{\algorithmicrequire}{\textbf{Input:}}
\renewcommand{\algorithmicensure}{\textbf{Output:}}
\begin{algorithm}
\caption{Greedy Coordinate Pairs (GCP)}
\label{alg:greedy-ascent}
\begin{algorithmic}
\REQUIRE Plot utility $U$, target point $\xvec_0$, max number of features $D$, grid size $M$
\ENSURE Optimized direction $\vvec^*$
\STATE $G(\ii, \ij, \theta) \equiv$ Givens rotation matrix for coordinates $\ii$ and $\ij$ with angle $\theta$
\STATE $\Theta = \{0, \frac{\pi}{M}, \frac{2\pi}{M}, \cdots, \pi\}$
\STATE $\vvec \gets \argmax_{\ii \in \{1,\dots, d\}} U(\tilde{f}_{\mathbf{e}_\ii})$ 
\COMMENT{Coordinate-wise optimization}
\WHILE{$\vvec$ not converged}
\STATE $\begin{aligned}
    \ii^*, \ij^*, \theta^* \gets &\textstyle{\argmax_{\ii, \ij, \theta \in \Theta}} U(\tilde{f}_{G(\ii, \ij, \theta)\vvec}) \\
    &\text{s.t.}  \quad {\textstyle \sum_k} \I([G(\ii, \ij, \theta)\vvec]_k \neq 0) \leq D
\end{aligned}$
\STATE $\vvec \gets G(\ii^*, \ij^*, \theta^*)\vvec$
\ENDWHILE
\end{algorithmic}
\end{algorithm}

\paragraph{Computational complexity.}
The total complexity of the GCP algorithm---which is easily parallelized---is $O(dkMI)$, where $M$ is the number of angles tested and $I$ is the number of iterations. In our experiments, GCP typically found an interesting direction within 10 iterations. 
By comparison, the computational complexity of standard PDP plots for all dimensions in terms of the number of model evaluations is $O(ndk)$. Thus, compared to PDP plots, our method is at most $O(MI)$ slower.

\begin{figure*}[!t]
\begin{subfigure}[b]{\textwidth}
  \centering
  \includegraphics[width=\linewidth]{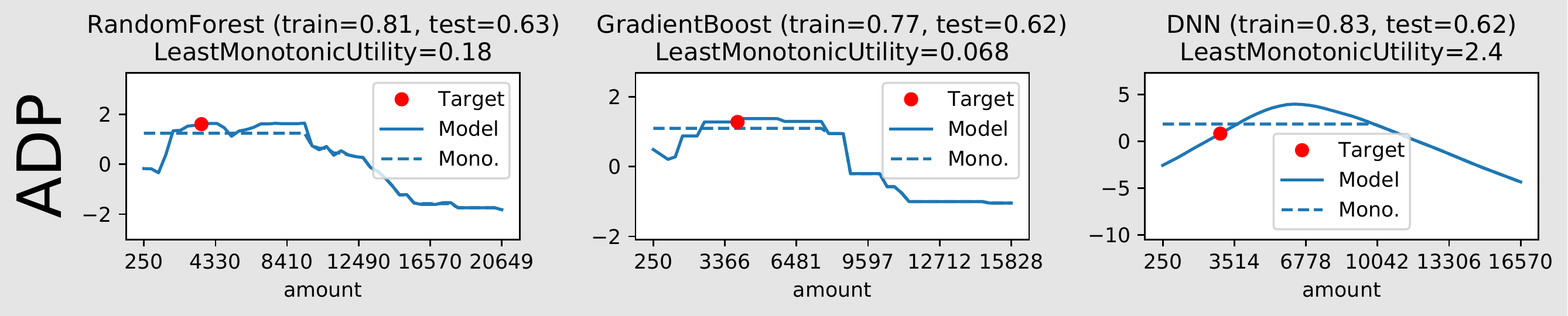}
  \label{fig:our_model}
\end{subfigure}
\begin{subfigure}[b]{\textwidth}
  \centering
  \vspace{-1em}
  \includegraphics[width=\linewidth]{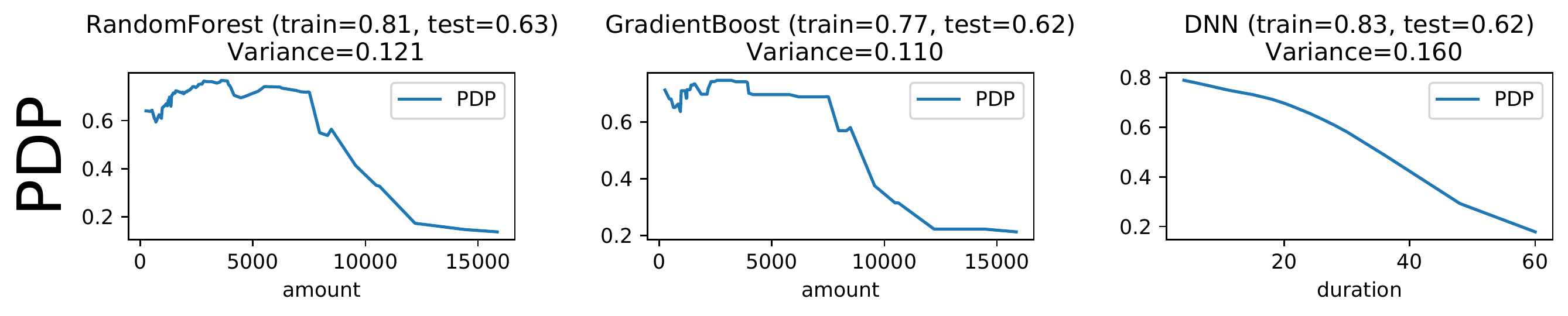}
  \label{fig:pdp}
\end{subfigure}
\begin{subfigure}[b]{\textwidth}
  \centering
  \vspace{-1em}
  \includegraphics[width=\linewidth]{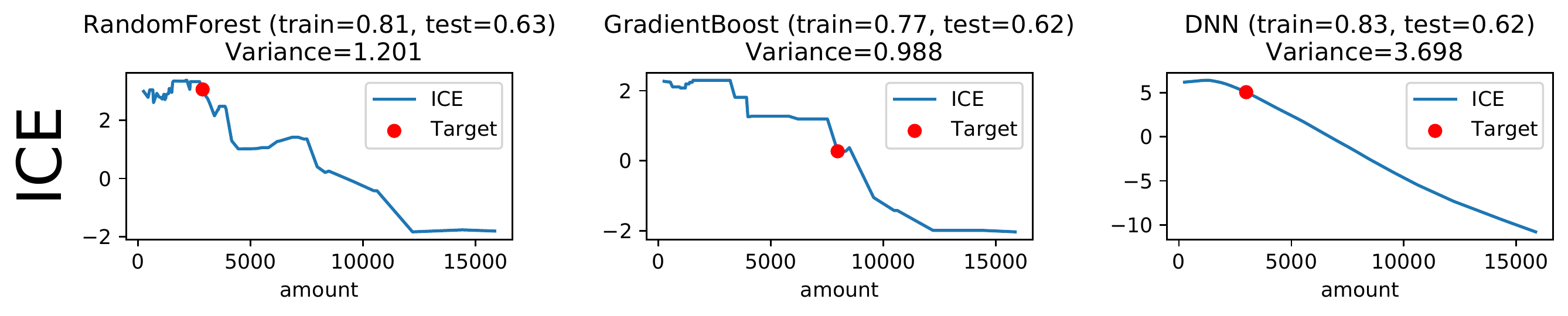}
  \label{fig:ice}
\end{subfigure}
\vspace{-2.5em}
\caption{
Our dependence plots (top, gray) selected by finding the least monotonic direction elucidates potentially problematic behavior of the various models whereas PDP (middle) and instance-specific ICE (bottom) plots do not highlight this non-monotonic behavior.
The PDP and ICE plots were selected according to the simple highest variance utility, which is used in practice for standard PDP plots.
All plots are based on 100 random samples, where the PDP plots average over the samples and our plots and the ICE plots additionally select a specific target point.
The dotted line (labeled ``Mono.'') is the best monotonic regression plot.
}
\label{fig:loan-monotonic-comparison}
\end{figure*}

\paragraph{Optimization evaluation.}
To test the effectiveness of GCP (Algorithm~\ref{alg:greedy-ascent}), we ran two tests.
First, we compared the optimal utility values returned by GCP to the utility of 10,000 randomly selected directional plots based on a random forest classifier.
In all cases, GCP returned the highest values.
For example, GCP found a directional plot $\tilde{f}$ with $U(\tilde{f})=0.0016$, compared with $0.0007$ for the best random curve.
In the second experiment, we generated random synthetic models in which the directional plot with the highst utility could be determined in advance, making it possible to evaluate how often GCP selects the optimal one.
We evaluated its performance on directions with at most one, two, or three nonzero coordinates, and found that in 100\%, 97\%, and 98\% of simulations, GCP found the optimal directions.
In the few cases GCP did not find the optimal directions, this was due to the randomness in generating examples whose optimal directions are more difficult to identify (e.g.,, the optimal curve was nearly constant).
Details and further results on these experiments can be found in the appendix.
Thus, we conclude that despite our algorithm being greedy, our GCP algorithm is empirically reasonable.

\section{EXPERIMENTS}
\label{sec:expt}

We present five concrete use cases: 1) Selecting among several models, 2) Bias detection, 3) Out-of-sample behavior in computer vision, 4) Discovering interesting model properties via generative models, and 5) Evaluating robustness to covariate shift.
We have put (5) and some details of the experiments in the appendix. A python module implementing the proposed framework is available at \url{https://github.com/davidinouye/adp}.

\subsection{SELECTING A MODEL FOR LOAN PREDICTION}
Suppose we have trained multiple models, each with similar test accuracies.
Which of these models should be deployed?
To answer this question, directional dependence plots can be used to detect undesirable or unexpected behaviours.
We explore this use-case of qualitative model selection on a dataset of German loan application data in order to find non-monotonic plots.
For example, it may be undesirable to deploy a model that decreases a candidate's score in response to an increase in their income.
We train a random forest, a gradient boosted tree, and a deep neural network. For this simple example, we consider axis-aligned directions with only one non-zero so that we can compare to PDP and ICE plots; additionally, we optimize the utility over 100 random target points $\xvec_0$ as in \autoref{eqn:main-optimization}---thus providing an estimate of the worst-case behavior of the model.
The test accuracies of these models were all close to 62\%---thus, a stakeholder would not be able to select a model based on accuracy alone.
The plots for our method using the monotonicity utility compared to PDP and ICE using the simple variance utility can be seen in \autoref{fig:loan-monotonic-comparison}.
In addition to a single number that quantifies non-monotonicity, our directional dependence plots show the user both the \emph{location} and \emph{severity} of the worst-case non-monotonicity.
On the other hand, PDP and instance-specific PDP (ICE) plots selected by variance prefer models that have large ranges, but may have expected or uninteresting patterns within this range. By contrast, our dependence plots can highlight more subtle and interesting patterns such as non-monotonicity.
Our directional dependence plots suggest that random forests or gradient boosted trees may be preferable since their worst-case behaviors are nearly monotonic, whereas the DNN model is far from monotonic.

\begin{figure}[!ht]
    \centering
    \includegraphics[width=\linewidth]{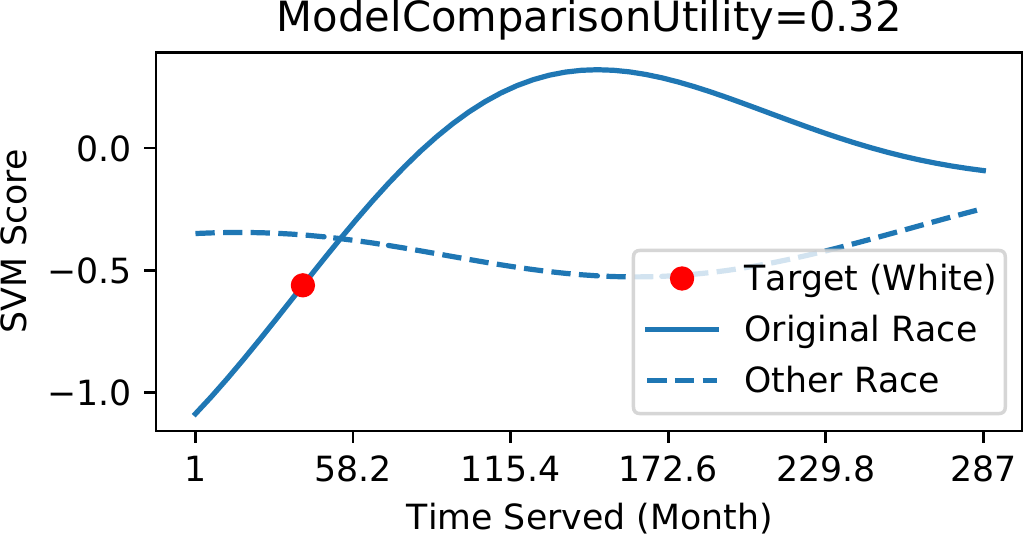}
    \vspace{-0.5em}
    \caption{
    A directional plot using the model contrast utility where the comparison model is the same model but the race is flipped.
    Notice that bias between races is far from uniform even switching bias direction.
    Expanded figure with flipping both gender and race on two target points can be found in the appendix.
    }
    \vspace{-1.5em}
    \label{fig:bias}
\end{figure}

\subsection{BIAS DETECTION IN RECIDIVISM MODEL PREDICTION}
In many real-world applications, it is essential to check models for bias.
A contemporary example of significant interest in recent years concerns \emph{recidivism prediction instruments} (RPIs), which use data to predict the likelihood of a convicted criminal to re-offend \citep{chouldechova2017fair,chouldechova2017fairer}.
Given an instance $\xvec_0=(u,x_{2},\ldots,x_{d})$, consider what the output of $f$ would have been had the protected attribute $u\in\{0,1\}$ (e.g., race or gender) been flipped. In certain cases, such protected attributes might not be explicitly available, in which case, we could use proxy attributes or factors, though we do not focus on that here. 
A model that ignores the protected attributes would see little or no change in $f$ as a result of this change.
In this simple example, we explicitly ignore dependencies between the protected attribute and other features though this would be important to consider for any significant conclusions to be made.
Given this situation, we select the model contrast utility (Sec~\ref{sec:utility}) with a special definition for the comparison model defined as follows:
$g_{\uvec_0}(\uvec) = f_{\uvec_0}(\sigma(\uvec)),$ where $[\sigma(\uvec)]_\ii$ is $1-u_\ii$ if the $\ii$-th feature is the protected attribute, and $u_\ii$ otherwise, essentially flipping only the protected attribute and leaving all other features untouched.
There are two cases: Either (a) No such plot deviates too much, in which case this is evidence that $f$ is not biased, or (b) there is a dimension along which $f$ is significantly biased.
A directional plot for flipping race from white to black based on data from \citep{schmidt1988predicting} using a kernel SVM model can be seen in Fig~\ref{fig:bias}.
One can clearly see that the model behaves in biased ways with respect to race: The effect of time served on the risk score clearly depends on race and even switches bias direction.
For this high stakes application, the use of such a model could be problematic, and our directional plots highlight this fact easily even for non-technical users.
Finally, these directional plots  avoid averaging the data into a single metric and offer more insight into the location and form of the model bias that may depend on the inputs in complex ways.

\subsection{UNDERSTANDING OUT-OF-SAMPLE BEHAVIOR FOR TRAFFIC SIGN DETECTION}
\begin{figure*}[!ht]
    \centering
    \includegraphics[width=\textwidth]{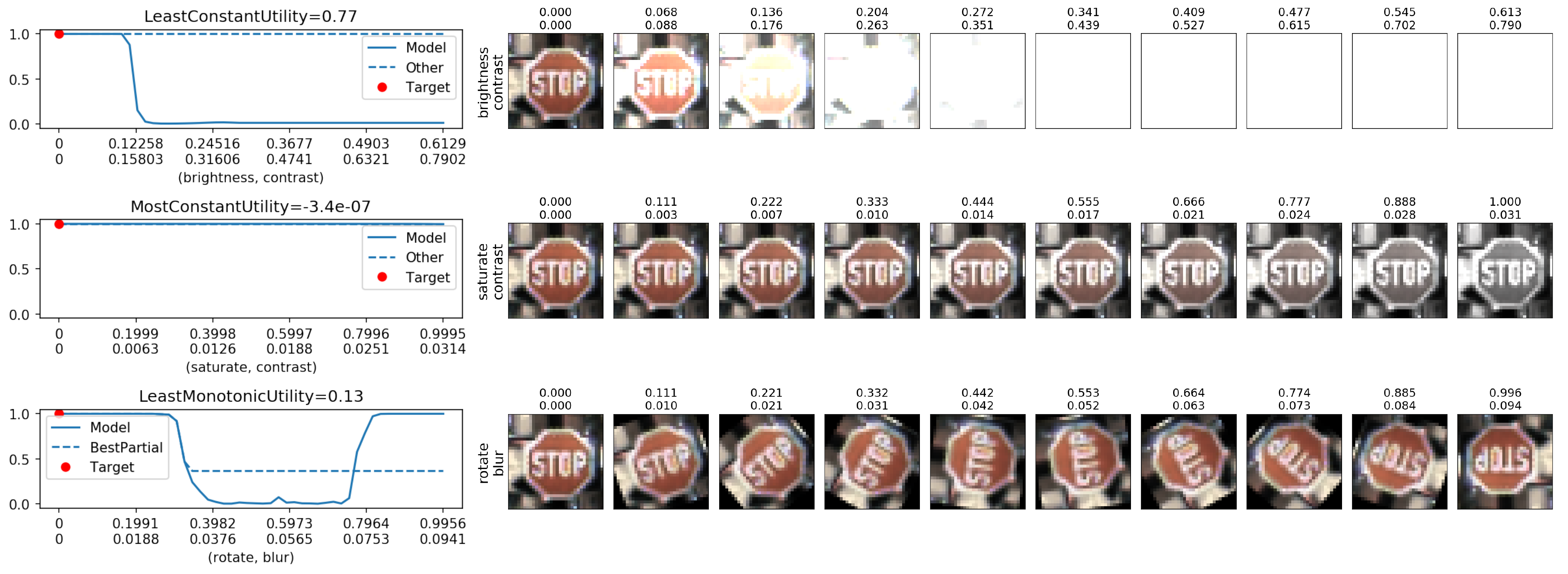}
    \vspace{-2em}
    \caption{
    The dependence plots 
    for traffic sign detection show which transformation is the most sensitive (top), the least sensitive (middle), and the least monotonic (bottom).
    These curves highlight both expected model trends (top) and unexpected trends (middle and bottom), where the model seems to ignore color and fails when the stop is rotated partially but works again when the stop sign is almost flipped over. 
    }
    \vspace{-1em}
    \label{fig:img-global-curve}
\end{figure*}

When the model is deployed in practice, understanding how the model behaves outside of the training dataset is often critical. 
For example, for a traffic sign detector on autonomous vehicles, will the detector predict correctly if a traffic sign is rotated even though the training data did not contain any rotated signs? What other variations of the data is the detector susceptible to?
For this use-case, we first trained a convolution neural network on German Traffic Sign Recognition Benchmark (GTSRB) dataset \cite{Stallkamp-IJCNN-2011} which achieves 97\% test accuracy, that will act as the stop sign detector.
We consider transformation curves based on five standard image transformations:
rotate, blur, brighten, desaturate, and contrast.
Each of these transformations creates images outside of the training data, and we are interested in finding the interesting combinations of these transformations that influence the prediction score of the detector in different ways.
\autoref{fig:img-global-curve} depicts the resulting plots we generated through optimizing the corresponding utilities.
The least constant direction (top) simultaneously adjusts contrast and brightness, which is expected since this transformation gradually removes almost all information about the stop sign.
The most constant direction (middle) adjusts saturation and contrast, which may be surprising since it suggests that the model ignores color.
Finally, the least monotonic direction (bottom) is rotation, which suggests that the model identifies a white horizontal region but ignores the actual letters ``STOP'' since it still predicts correctly when the stop sign is flipped over.

\begin{figure*}[!ht]
    \centering
    \includegraphics[width=\textwidth]{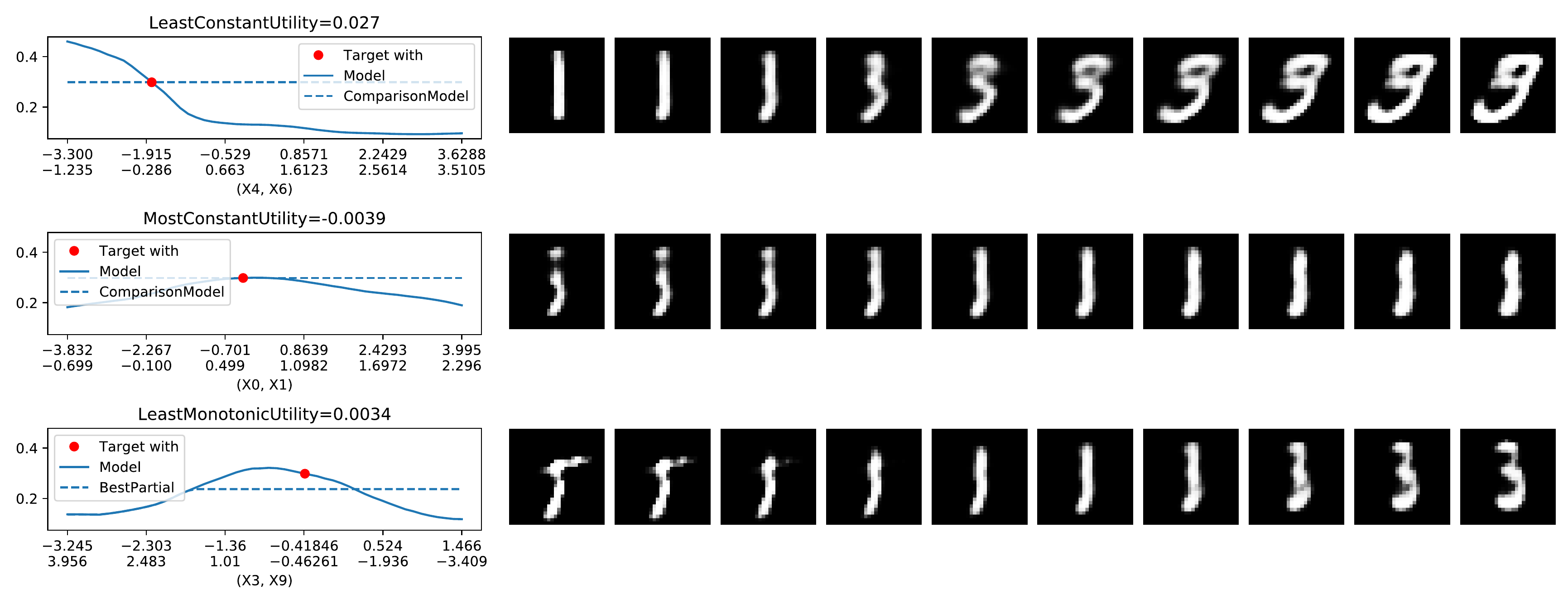}
    \vspace{-1em}
    \caption{Discovering various non-linear directions in images using VAE. Given the target instance of digit 1, the least-constant utility highlights that adding more curves and loops to the image induces the prediction probability to drop the most (top row), while the most-constant utility identifies directions that preserve the relative shapes of the digit (middle row). The least-monotonic utility exposes the directions that make the prediction probability fluctuate the most -- moving along these directions alter the digit to different numbers, e.g., 5 and 3.}
    \label{fig:vae}
\end{figure*}

\subsection{DISCOVERING INTERESTING MODEL PROPERTIES VIA GENERATIVE MODELS}

Our directional dependence plots can also search non-linear directions (i.e., curves) in the input space, by discovering linear directions in some latent feature space learned by generative models. This can be particularly useful when linear directions in the input feature space are less semantic compared to non-linear directions, like images. In this example, we demonstrate how directional plots effectively select the most interesting non-linear directions in the pixel space. 

On the MNIST data, we train a VAE~\cite{Kingma2014} for learning a 10-dimensional latent data representation, as well as a CNN classifier. The directions in the latent space of deep networks are found to be semantic in several ways~\cite{kim2018interpretability}, so we optimize in this space for the least constant, most constant, and least monotonic in terms of the prediction score change of the classifier. In Figure~\ref{fig:vae},  given a test instance of digit 1, we show the varying prediction probability (left) along with the set of reconstructed images along the direction (right). 
Least-constant utility (top row) discovers directions that add distinctive curves and loops to the image inducing a steep drop in the prediction probability, while the most-constant utility (middle row) finds directions that keep the relative shape of the digit 1 constant throughout. 
Least-monotonic utility (bottom row) successfully exposes directions that change the digit 1 to 5 and 3, inducing the most non-monotonic prediction probability changes. 
Note that such discovery is not limited to images --- as long as some semantically meaningful feature embedding is available (e.g., word embedding), the same process can be used to automatically discover interesting directions within the input feature space.
While these use cases demonstrate our method across a wide range of scenarios, a user study for various contexts such as exploratory data analysis, model checking, or model comparison would be an excellent area for future research.
Overall, we believe our framework opens up new possibilities and challenges for automating and extending PDP plots.

\subsubsection*{Acknowledgements}
The authors acknowledge the support of DARPA via FA87501720152 and Accenture.
DII acknowledges support from Northrop Grumman.
JSK acknowledges support from Kwanjeong Educational Foundation.
BA acknowledges support from the Robert H. Topel Faculty Research Fund.
 
\subsubsection*{References}
{
\renewcommand{\section}[2]{}
\footnotesize
\bibliographystyle{abbrvnat}
\bibliography{ref,other,Mendeley-fixed}
}

\clearpage
\appendix

\section{Evaluating robustness under covariate shift in loan prediction}
\begin{figure}[!ht]
     \centering
     \includegraphics[width=0.8\linewidth]{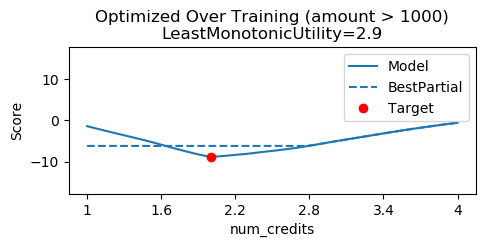}
     \includegraphics[width=0.8\linewidth]{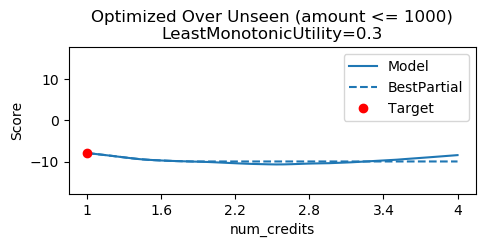}
     \vspace{-1em}
     \caption{
     Our dependence plots selected by finding the least monotonic direction for deep neural network model trained on large loan applications (amount > 1,000) when optimizing over (top) the training distribution (i.e., amount > 1,000) and over (bottom) the unseen novel small loan distribution (i.e., amount $\leq$ 1,000).
     The dotted line (labeled ``BestPartial'') is the best isotonic regression model.
     }
     \label{fig:domain-loan}
     \vspace{-1.5em}
 \end{figure}

Using the same dataset as in the previous example, we compare the behavior of the \emph{same} model over different regions of the input space. The motivation is covariate shift: A bank has successfully deployed a loan prediction model but has historically only focused on high-income customers, and is now interested in deploying this model on low-income customers. Is the model robust? Once again we use dependence plots to detect undesirable behaviour such as non-monotonicity. We trained a deep neural network that achieves 80\% accuracy on high-income data and a comparable 76\% accuracy on the unseen low-income data. Moving beyond accuracy, we generated directional plots using the least monotonic utility optimizing over all target points in the training data (i.e., high-income) and the unseen test data (i.e., low-income) as can be seen in \autoref{fig:domain-loan}. The curves explicitly display the  difference between the worst-case non-monotonicity for high-income (\autoref{fig:domain-loan}, top) and low-income (\autoref{fig:domain-loan}, bottom), which appears to be minimal, giving stakeholders confidence for deploying this model.

\section{Lipschitz-bounded property validation formulated as a constrained least squares problem}
Suppose we have a grid of $t_i$ values and corresponding to model outputs along the curve $y_i=f(\curve(t_i))$.
Now let $\bm{b}=\yvec$ and let $A$ be defined as follows:
\begin{align}
    A = \begin{bmatrix}
    1 & 0 & 0 & 0 & 0 \\
    1 & t_2 - t_1 & 0 & 0 & 0 \\
    1 & t_2 - t_1 & t_3 - t_2 & 0  & 0\\
    \vdots & \vdots & \ddots &  \ddots & \vdots \\
    1 & t_2 - t_1 & \dots & \dots & t_{n} - t_{n-1}
    \end{bmatrix} \, .
\end{align}
Now we solve the following simple least squares problem:
\begin{align}
\begin{aligned}
    \hat{\xvec} = \argmin_{\xvec} &\quad \|A\xvec - \bm{b}\|^2_2 \\
    \textnormal{s.t.}  & -L \leq x_i \leq L, \quad \forall i \in \{2,n\} \,.
\end{aligned}
\end{align}
Notice that the first coordinate $x_1$ is unconstrained and represents $\hat{y}_1$.
The rest of $\xvec$ correspond to the slope of a line connecting each point; thus $\hat{y}_2 = \hat{y}_1 + (t_2 - t_1)x_2$ and $\hat{y}_3 = \hat{y}_1 + (t_2 - t_1)x_2 + (t_3 - t_2)x_3$, etc.
Thus, $\hat{\yvec} = A\hat{\xvec}$ and our approximation is merely a linear interpolation using $\bm{t}$ and $\hat{\yvec}$.

\section{Synthetic optimization figure}
\begin{figure*}[!ht]
    \centering
    \includegraphics[width=\textwidth]{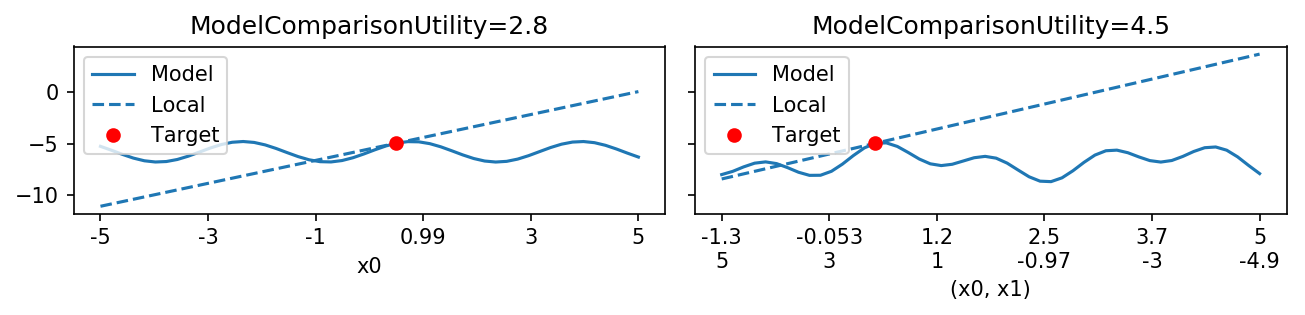}
    \includegraphics[width=\textwidth]{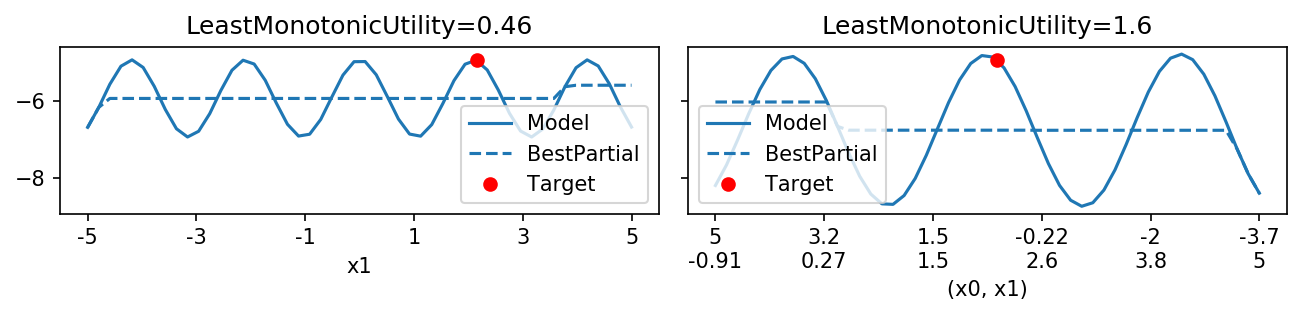}
    \caption{Our dependence plots when optimizing the synthetic function defined in \ref{app:opt} using the model comparison utility to a first-order Taylor series approximation (top) and using the least monotonic utility (bottom).
    In both cases, our optimization algorithm indeed finds the correct direction along $x_0$ and $x_1$.
    }
    \label{fig:synthetic}
\end{figure*}

See \autoref{fig:synthetic}.

\section{Expanded figure for bias detection}
\begin{figure*}[!ht]
\centering
\begin{subfigure}[b]{0.48\textwidth}
  \centering
  \includegraphics[width=\linewidth]{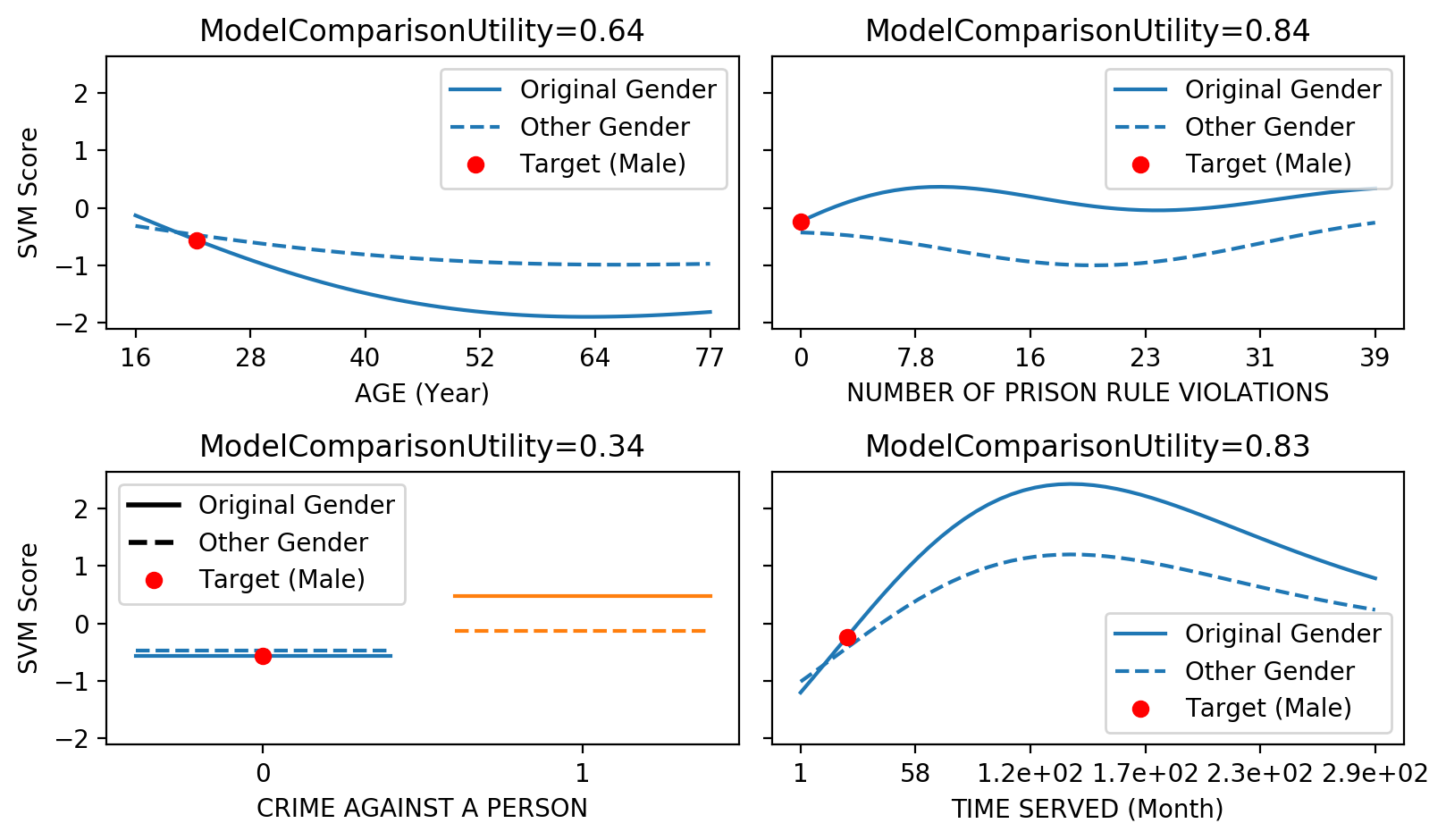}
  \caption{Flipping gender}
  \label{fig:flip-gender}
\end{subfigure}
\begin{subfigure}[b]{0.48\textwidth}
  \centering
  \includegraphics[width=\linewidth]{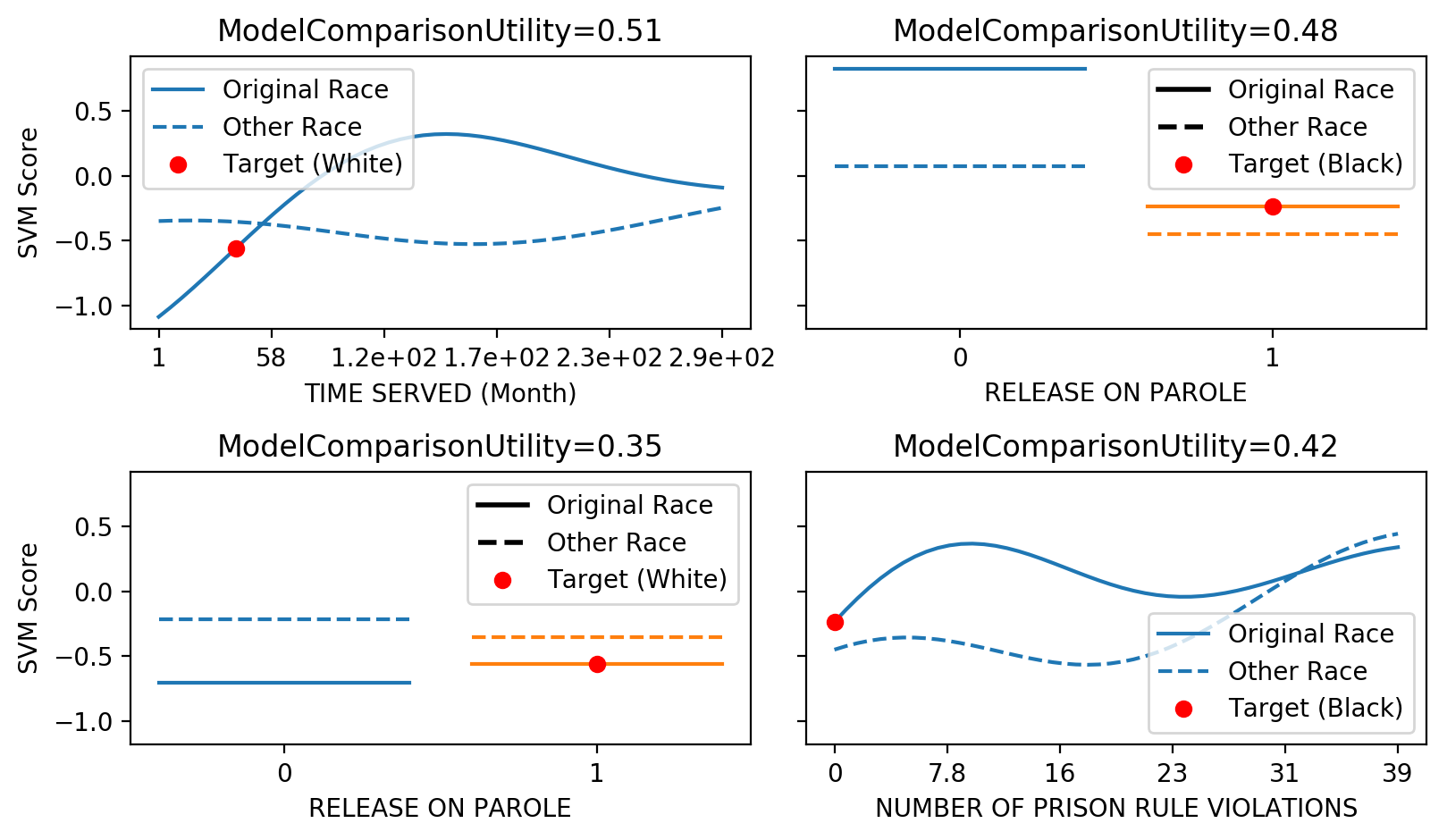}
  \caption{Flipping race}
  \label{fig:flip-race}
\end{subfigure}
\caption{Our dependence plots showing top two biased features (rows) of two target instances (columns) for flipping (a) gender  and (b) race.
Notice that bias between groups is quite evident and is far from uniform; sometimes the bias even switches depending on the feature values (top left of subfigure (b)).
    }
    \label{fig:img-race-gender}
\end{figure*}

See \autoref{fig:img-race-gender}.

\section{Optimization evaluation details}
\label{app:opt}

\paragraph{Synthetic experiment.}

We create a synthetic model $f$ to test our optimization algorithm.
Consider a function $f(\xvec) = \sin (2x_0) + \cos (3x_1) + \beta^T\xvec_{2:}$ for $x_0, x_1 \in \mathbb{R}$, and $\beta, \xvec_{2:} \in  \mathbb{R}^{\ndim-2}$.
We consider two utilities: 1) the model comparison utility with the first-order Taylor series approximation to $f$, i.e., $g_{\xvec_0}(\xvec) \equiv f(\xvec_0) + \nabla f(\xvec_0)^T \xvec$, and 2) the least monotonic utility.
It can be seen that the ground-truth best linear curve, for the model $f$ above, with respect to these two utilities, will have directions along $x_0$ and $x_1$ respectively. Sinusoidal functions are indeed less monotonic than linear functions, and they also deviate away from the first-order Taylor series approximation, which is also linear.  We verify that these directions are correctly found using our optimization algorithm as seen in Figure~\ref{fig:synthetic}.

To more generally verify that GCP finds correct direction that maximizes the least monotonic utility, we simulated random model behaviors that are non-monotonic in certain directions, and compared the direction found with GCP with the ground-truth curves. For introducing non-monotonicity, a random set of polynomial functions were used, additionally constraining that the utilities of the curves along the ground-truth directions are non-zero and above certain threshold, while the utilities along the non-ground-truth directions are relatively small or almost zero.

\begin{figure}[t]
\vspace{-1em}
    \centering
    \includegraphics[width=0.75\linewidth]{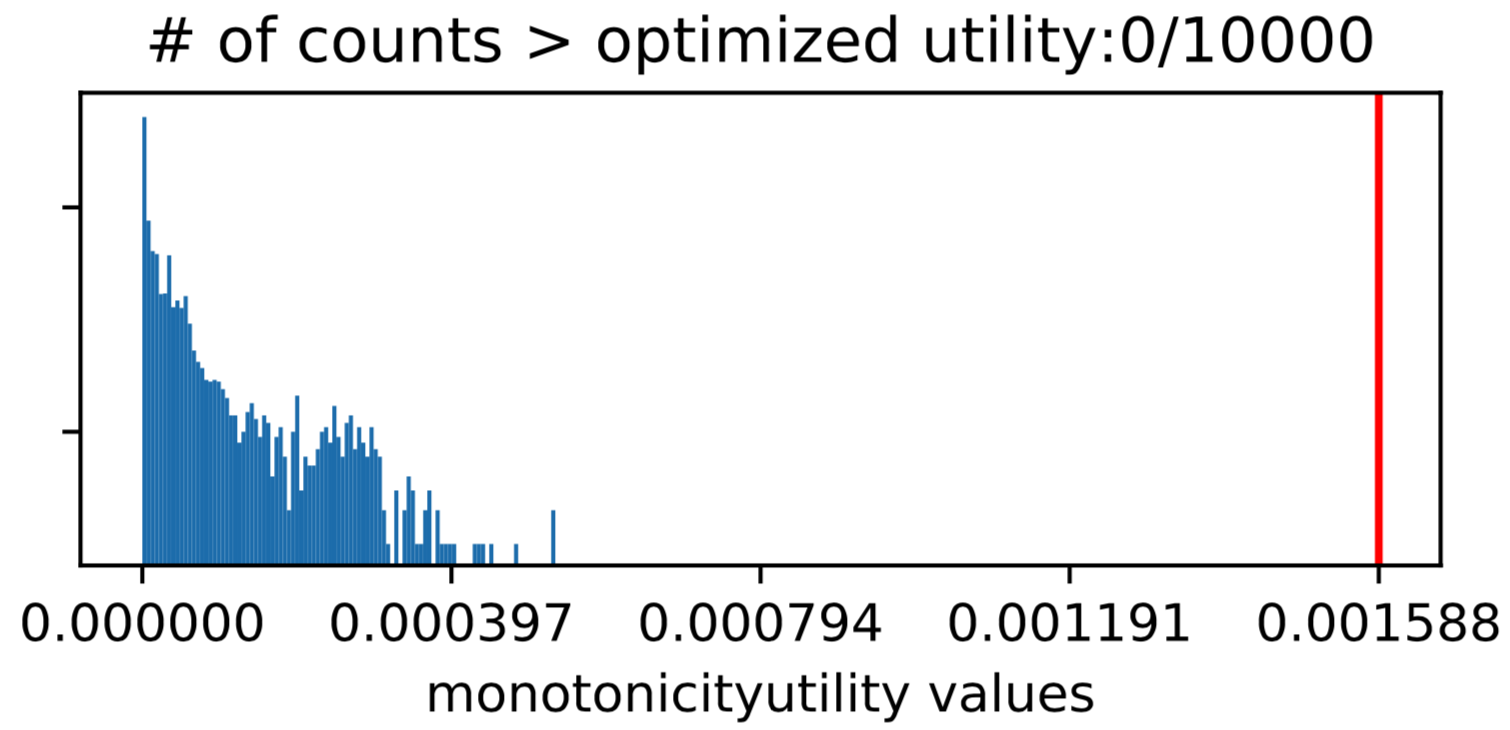}
    \caption{The utility found by our optimization (red line) is clearly finding a large value for utility compared to random directions (blue histogram with counts in log scale) demonstrating that our optimization method performs well empirically.
    }
    \label{fig:utility-histogram}
\end{figure}

\paragraph{Utility histogram.}
Another empirical way to check our optimization method is to randomly sample curves and compute their utilities; then we can compare to the utility of our optimized curve.
We generate directions using a random forest on the loan application data (see 
\autoref{sec:expt} for more data details).
For interpretability, we restrict the curve parameter $\vvec$ to only have three non-zeros.
We sample uniformly from all directions that have at most three non-zeros, i.e., $\vvec \in \{ \vvec \in \R^\ndim: \sum_i \one(v_i \neq 0) \leq 3 \}$.
We can see in the histogram of utility values (log scale) shown in \autoref{fig:utility-histogram} that the utility of our optimized curve (red line) is clearly better than the utility of random directions.
In addition, we note that even if we do not find the global optimum, our optimized diagnostic curves can still be useful in highlighting interesting parts of the model---see use-case experiments.
Thus, This shows that even though the optimization problem is quite difficult, we can perform well empirically.

\begin{table*}
\caption{Parameter Values for Models}
\label{tab:parameters}
\begin{tabular}{lp{0.7\textwidth}}
    Model Name & Parameters \\
    \toprule
    Decision tree & Max leaf nodes $\in \{5, 10, 20, 40\}$, max depth $\in \{1, 2, \cdots, 10\}$ \\
    Gradient boosted trees & Learning rate $\in \{0.05, 0.1, 0.2\}$, Number of estimators $\in \{25, 50, 100, 200, 500\}$ \\
    Deep NN & Max epoch =  1000, Learning rate = 0.0001, batch size $\in \{100, 200, 400\}$, two hidden layers of size 128 with relu activations and softmax final activation, ADAM optimizer and BCE loss.\\
    \bottomrule
\end{tabular}
\end{table*}

\section{More experiment details}
\paragraph{Selecting model for loan prediction.}
The data used in this experiment is German loan application data.\footnote{\url{https://archive.ics.uci.edu/ml/datasets/statlog+(german+credit+data)}} This dataset has 7 numeric attributes and 13 categorical attributes ranging from the amount of the loan requested to the status of the applicant's checking account.

We train a random forest, a gradient boosted tree, and a deep neural network on only the numeric attributes (since monotonicity isn't well-defined for categorical attributes).
We tune each model via cross validation using \texttt{scikit-learn}.
We optimize each model over the parameters in \autoref{tab:parameters} (where other parameters are defaults in \texttt{scikit-learn}).

\paragraph{Evaluating robustness under covariate shift in loan prediction.}
To simulate this setup, we split the German loan dataset based on amount: a training dataset with 884 users with \texttt{amount} $>$ 1,000 DMR and a separate unseen test dataset of 116 users with \texttt{amount} $\leq$ 1,000 DMR---note that these will give two different data distributions.
We train via the deep neural network parameters and cross validation in \autoref{tab:parameters}.

\paragraph{Understanding out-of-sample behavior for traffic sign detection.}
We train a convolution neural network on German Traffic Sign Recognition Benchmark (GTSRB) dataset \cite{Stallkamp-IJCNN-2011} which achieves 97\% test accuracy. 
We consider directions 
based on five image transformations:
rotate, blur, brighten, desaturate, and increase contrast.
Each image transformation will create images outside of the training distribution---hence, we can view the behavior of the model outside of the training data.

 \end{document}